\documentclass[runningheads]{llncs}

\usepackage[final,year=2024,ID=9881]{eccv}

\usepackage{eccvabbrv}

\usepackage{graphicx}
\usepackage{booktabs}

\usepackage[accsupp]{axessibility}  

\usepackage[pagebackref,breaklinks,colorlinks,citecolor=eccvblue]{hyperref}

\usepackage{orcidlink}

\begin{document}

\title{MeshSegmenter: Zero-Shot Mesh Semantic Segmentation via Texture Synthesis} 

\titlerunning{MeshSegmenter}


\author{Ziming Zhong\inst{1} \and
Yanyu Xu\inst{2} \and
Jing Li\inst{3}\and
Jiale Xu\inst{1} \and
Zhengxin Li\inst{1} \and
Chaohui Yu\inst{4} \and
Shenghua Gao\inst{1,5}}

\authorrunning{Z. Zhong, Y. Xu et al.}


\institute{\mbox{ShanghaiTech University \and Shandong University \and Xiaohongshu Inc \and Alibaba Group}   \and The University of Hong Kong \\
\email{\{zhongzm, lijing1, xujl1, lizhx, gaoshh\}@shanghaitech.edu.cn} 
\mbox{ \email{xu\_yanyu@sdu.edu.cn} \qquad \email{huakun.ych@alibaba-inc.com}}}


\renewcommand\twocolumn[1][]{#1}
\maketitle
\begin{center}
\centering
\includegraphics[width=1.0\linewidth]{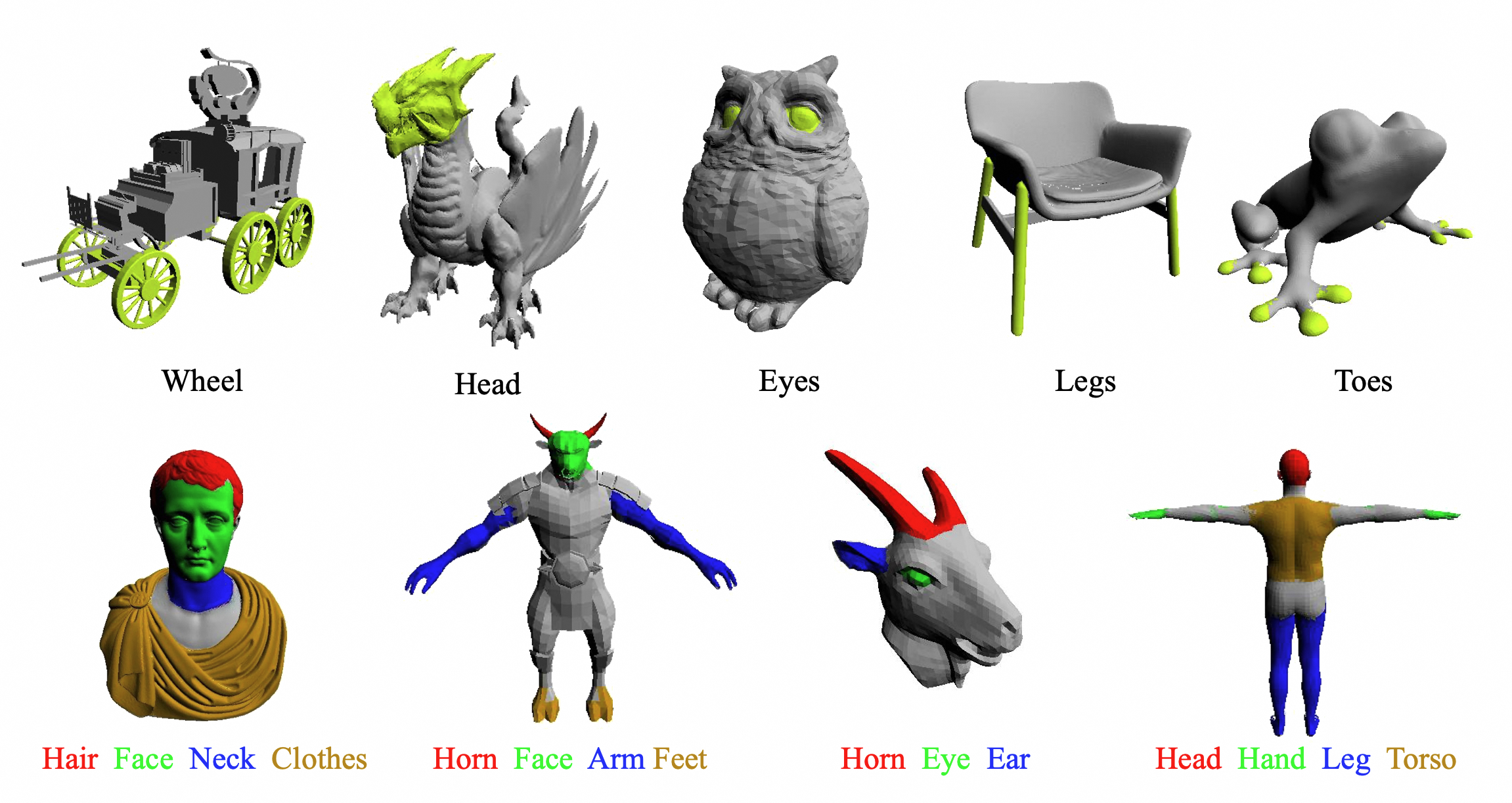}
\captionof{figure}{MeshSegmenter performs zero-shot mesh semantic segmentation through texture synthesis. It can accurately segment the text-specified region by aggregating multi-view 2D segmentation results from single and multiple queries.}

 \label{fig:teaser}
\end{center}%

\begin{abstract}
We present MeshSegmenter, a simple yet effective framework designed for zero-shot 3D semantic segmentation. This model successfully extends the powerful capabilities of 2D segmentation models to 3D meshes, delivering accurate 3D segmentation across diverse meshes and segment descriptions. Specifically, our model leverages the Segment Anything Model (SAM) model to segment the target regions from images rendered from the 3D shape. In light of the importance of the texture for segmentation, we also leverage the pretrained stable diffusion model to generate images with textures from 3D shape, and leverage SAM to segment the target regions from images with textures. Textures supplement the shape for segmentation and facilitate accurate 3D segmentation even in geometrically non-prominent areas, such as segmenting a car door within a car mesh. To achieve the 3D segments, we render 2D images from different views and conduct segmentation for both textured and untextured images. Lastly, we develop a multi-view revoting scheme that integrates 2D segmentation results and confidence scores from various views onto the 3D mesh, ensuring the 3D consistency of segmentation results and eliminating inaccuracies from specific perspectives. Through these innovations, MeshSegmenter offers stable and reliable 3D segmentation results both quantitatively and qualitatively, highlighting its potential as a transformative tool in the field of 3D zero-shot segmentation. The code is available at \url{https://github.com/zimingzhong/MeshSegmenter}.


\end{abstract}

\section{Introduction}
\label{sec:intro}

Segmenting semantic regions within a 3D mesh is crucial in the fields of computer graphics and computer vision, as highlighted in works such as \cite{chen2009benchmark, michele2021generative, decatur20223d}. However, the availability of accurately annotated 3D data is limited, and obtaining such data is a costly work. As a result, existing models trained on these data often struggle to generalize well to unseen examples. To mitigate this challenge, using open vocabularies as inputs for segmenting regions has emerged as an efficient approach, showcasing the practical value of 3D zero-shot mesh segmentation \cite{decatur20223d}. This task requires a model that has a comprehensive understanding of both the overall object and its local components. However, understanding local regions in 3D meshes can introduce ambiguities, leading to erroneous segmentations. For example, black clothing can be mistaken for hair when viewed from certain perspectives. Consequently, 3D zero-shot mesh segmentation represents a highly challenging task. In this paper, we propose a network framework specifically designed for zero-shot 3D semantic segmentation. Our approach leverages object descriptions and segmentation region descriptions to guide the segmentation process. By incorporating both descriptions, our model can automatically segment fine-grained semantic regions. To demonstrate the effectiveness of our method, we showcase its application to mesh segmentation and editing tasks.

Recently, large-scale 2D segmentation models~\cite{kirillov2023segment, liu2023grounding, Bansal_Sikka_Sharma_Chellappa_Divakaran_2018} have achieved remarkable success in various 2D segmentation tasks, showcasing exceptional zero-shot generalization capabilities. These achievements have been facilitated by extensive datasets of 2D images with corresponding annotations. However, acquiring equivalent annotated datasets for 3D data is a considerable challenge. Furthermore, the parameter size and computational costs associated with 2D segmentation models make them impractical for handling larger 3D datasets that demand higher quality segmentation. To address these limitations, we propose leveraging the powerful capabilities of 2D zero-shot models for 3D zero-shot segmentation, presenting a more efficient approach. Our proposed model initially performs zero-shot 2D segmentation on rendered mesh images. Subsequently, the segmentation results from different viewpoints are combined to generate a coherent 3D segmentation outcome. This approach ensures consistency in segmentation results across multiple viewpoints while integrating the results from neighboring viewpoints to enhance the robustness of the segmentation outcome.

Although current approaches to 3D zero-shot segmentation~\cite{decatur20223d, abdelreheem2023satr, abdelreheem2023zero} primarily rely on untextured meshes, the absence of texture information exacerbates the ambiguity of meshes across different viewpoints. However, recent advancements in generative models have enabled significant progress in generating multi-view consistent mesh textures by merging results from multiple viewpoints. By leveraging the geometric information obtained from these generative models, we can incorporate realistic texture information that helps reduce ambiguity in segmentation regions. Additionally, existing 2D segmentation models are trained on real images, creating a substantial domain gap when detecting untextured meshes, as shown in Fig.~\ref{fig:performance}. To address this issue, our model first generates high-quality textured meshes based on the untextured meshes and object mesh descriptions. This approach allows us to harness the potential of 2D segmentation models while benefiting from the purer geometric information provided by the untextured mesh model. Consequently, we perform 3D zero-shot segmentation on both untextured meshes and textured meshes, capitalizing on their respective strengths.

By adopting this novel approach, our proposed model addresses these challenges associated with untextured meshes. Through experimental evaluation, we demonstrate the efficacy and potential of our method for various 3D zero-shot segmentation tasks. Our work contributes to advancing 3D zero-shot segmentation and paves the way for further research in this domain.

Our contributions can be summarized as follows:
\begin{itemize}

\item We have introduced MeshSegmenter, a simple yet effective framework for 3D zero-shot semantic segmentation that efficiently elevates the capabilities of robust 2D segmentation models to 3D meshes. MeshSegmenter consistently presents precise 3D segmentation results across various meshes and a rich array of segment descriptions.

\item We have proposed the generation of textures based on object descriptions to enhance 2D segmentation models by providing additional texture information, thereby assisting these models in achieving accurate results. Leveraging latent texture information excavated from the generative models based on 3D mesh, our model can accurately perform 3D segmentation in geometrically non-prominent areas, such as segmenting the door in a car mesh.

\item We have developed a multi-view revoting module that integrates 2D detection results and confidence scores from various views onto the 3D mesh, effectively ensuring the 3D consistency of segmentation results and eliminating incorrect detection results from certain perspectives. This feature enables our model to provide stable and reliable 3D segmentation results.
\end{itemize}
\section{Related Work}


\begin{figure*}[t]
\centering
\includegraphics[width=\textwidth]{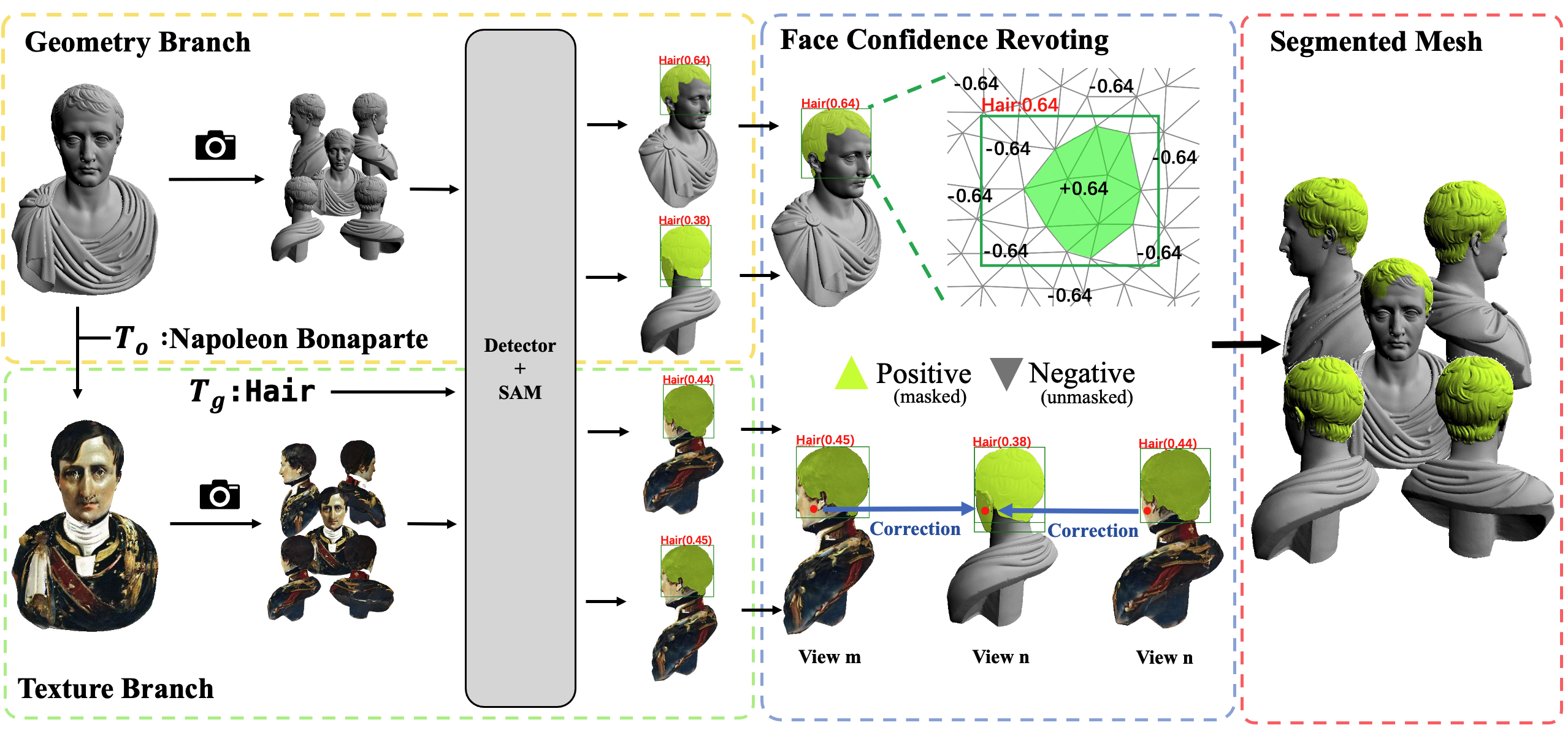}
\caption{Overview of the proposed pipeline. The Stable Diffusion (SD) model can generate high-quality textures under the guidance of textual prompts. Textured and untextured meshes are rendered from a fixed perspective. The rendered images are processed by GroundingDino\cite{liu2023grounding} and SAM\cite{kirillov2023segment}, which detect bounding boxes with corresponding confidence scores and segment the specific regions guided by these boxes, respectively. Ultimately, we employ face confidence revoting (FCR) to aggregate the detection and segmentation results of the textured and untextured meshes from multiple viewpoints to revote the 3D-awara scores to triangles.}
\label{fig:method}
\end{figure*}

\noindent\textbf{Zero-shot 2D Segmentation.}
Zero-shot 2D semantic segmentation is an important research area \cite{bansal2018zero, Rahman_Khan_Porikli_2018,Zhang_Ding_2021,Luddecke_Ecker_2022} and a challenging problem \cite{zou2022generalized, Xu_Zhang_Wei_Hu_Bai_2023, Xu_Mello_Liu_Byeon_Breuel_Kautz_Wang}, which could be divided into two groups.
The first group \cite{bucher2019zero, gu2020context, baek2021exploiting, pastore2021closer, zhang2021prototypical} enhances the generalization capability to unseen categories by aligning the semantic features of images with the image recognition model.
For instance, Bucher \emph{et al.} \cite{bucher2019zero} first introduced the task of zero-shot semantic segmentation, which aims to learn pixel-wise classifiers for never-seen object categories with zero training examples. 
In \cite{baek2021exploiting}, Baek \emph{et al.} propose to leverage visual and semantic encoders to learn a joint embedding space, where the semantic encoder transforms semantic features to semantic prototypes.
The second group \cite{ding2022decoupling, luddecke2022image, ghiasi2022scaling, xu2023side, xu2021simple} focuses on adapting rich semantic spaces of large-scale multi-modal neural networks, such as \cite{radford2021learning,mu2022slip,jia2021scaling} for segmentation tasks. 
These approaches enhance the generalizability to unseen categories through the combination of text, captioning, and self-supervision. 
Recent advancements \cite{xu2023learning, zhang2022glipv2,zou2022generalized} in zero-shot 2D semantic segmentation based on powerful large-scale multimodal models and natural language representations have shown significant improvement.
These approaches have demonstrated remarkable performance in generalizing to unseen categories.

\noindent\textbf{Zero-shot 3D Segmentation.}
In zero-shot 3D segmentation\cite{Koo_Huang_Achlioptas_Guibas_Sung, Lombardi_Simon_Saragih_Schwartz_Lehrmann_Sheikh_2019, Ding_Yang_Xue_Zhang_Bai_Qi_2022, Michele_Boulch_Puy_Bucher_Marlet_2021}, a notable focus has been placed on point cloud segmentation, as evidenced by pioneering works such as\cite{michele2021generative, chen2022zero,ding2022language,liu2022partslip} . The core idea is to enable models to semantically segment 3D shapes or point clouds without having seen examples of these specific categories during training. This research area leverages advancements in zero-shot learning from 2D image processing and applies them innovatively to 3D data.
Recent advancements\cite{Michele_Boulch_Puy_Bucher_Marlet_2021, kobayashi2022decomposing,shafiullah2022clip,goel2022interactive}in zero-shot 3D segmentation have been largely influenced by the integration of 2D image recognition models into 3D shape analysis. 
Recent advancements from Neural Radiance Fields (NeRFs) \cite{mildenhall2021nerf, lombardi2019neural} focus on constructing semantic fields. These approachs \cite{zhi2021place, vora2021nesf, fan2022nerf, kundu2022panoptic} enforce consistency between the rendered multi-view semantic labels and the ground truth or estimated semantic labels.


We categorize zero-shot segmentation based on its ability to segment a single or multiple queries into single query segmentation and multiple queries segmentation.

\noindent\textit{Single Query Segmentation.} In recent work on 3D zero-shot segmentation~\cite{mildenhall2021nerf, lombardi2019neural, decatur20223d}, the method known as 3DHighlighter \cite{decatur20223d} employs a neural network to predict the probabilities of mesh vertices belonging to the text-specified region. By coloring the vertices based on these probabilities and rendering the mesh from random viewpoints, the approach constrains the distance between the rendered images and the given text prompt in CLIP\cite{radford2021learning} space to obtain the segmentation results.

\noindent\textit{Multiple Queries Segmentation.} The recent advancements in visual-text pre-trained models have enabled the possibility of zero-shot segmentation of 3D objects into multiple parts. SATR\cite{abdelreheem2023satr} employs the GLIP\cite{Li_2022_CVPR} approach to obtain semantic segmentation results of images rendered from various perspectives, subsequently aggregating these multi-view segmentations. For each facet, the prompt with the highest confidence level is selected as the semantic segmentation result. Benefiting from the correction of low-confidence erroneous semantic segmentation by high-confidence accurate ones, SATR\cite{abdelreheem2023satr} achieves commendable performance. ZSC\cite{abdelreheem2023zero} uses ChatGPT to get all segmented prompts about the object and focuses on 3D shape correspondence in different objects. However, their reliance on the error-correcting capability of multi-queries semantic segmentation falters under the conditions of single-query segmentation.

\begin{figure}[!t]
\centering
\includegraphics[width=0.7\textwidth]{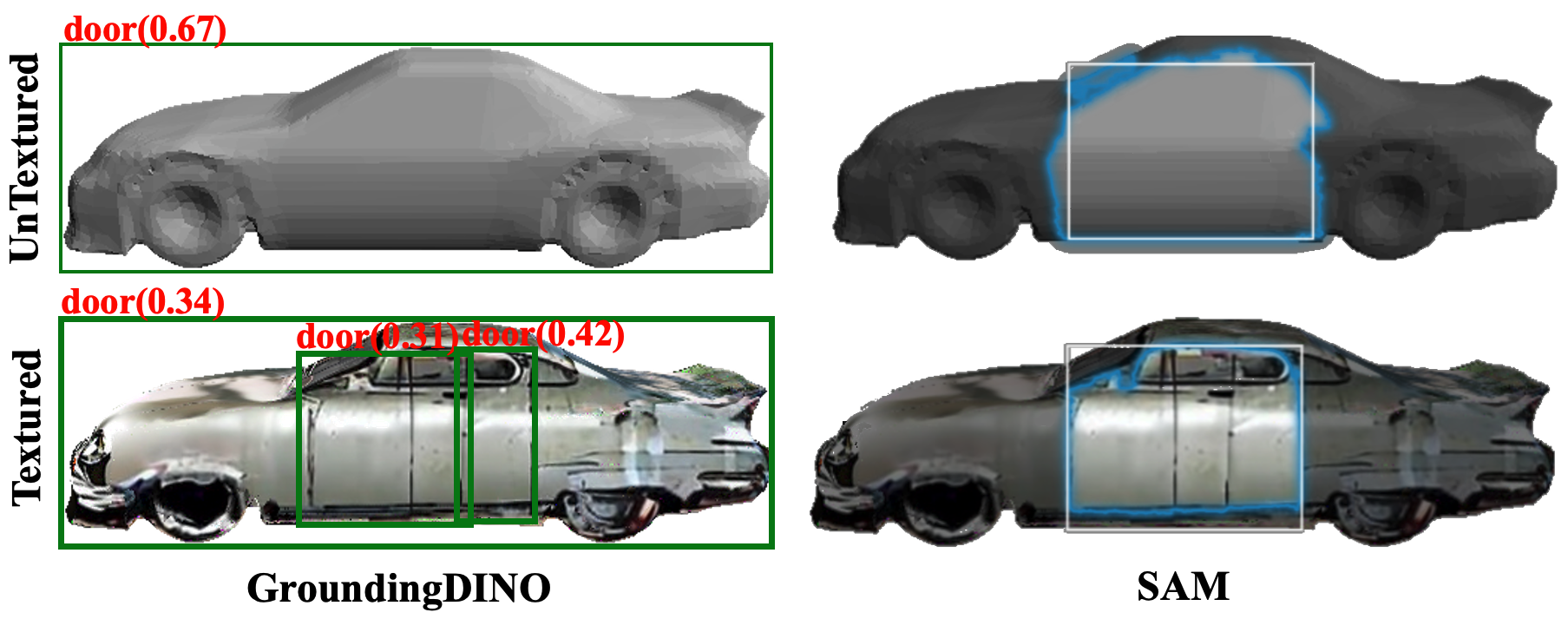}

\caption{
Performance of GroundingDINO\cite{liu2023grounding} and SAM\cite{kirillov2023segment} on $M_\text{Texture}$ and $M_\text{unTextured}$. The large-scale pre-trained models are typically trained on images rich in texture. However, there is a domain gap when these models are applied to  untextured meshes. GroundingDINO \cite{liu2023grounding} and SAM\cite{kirillov2023segment} are respectively provided with textual prompts and bounding boxes for detection and segmentation tasks, demonstrating significantly superior performance on textured images compared to performance on untextured images.
}

\label{fig:performance}
\end{figure}

\section{Method}

\subsection{Overview}

Mesh segmentation aims to segment the semantic regions according to the corresponding text descriptions. 
The input is a untextured mesh $M_\text{unTextured}$, characterized by vertices $V \in \mathbb{R}^{n\times 3}$ and faces $F \in {\{1,\cdots,n\}}^{m \times 3}$, where $m$ is the number of faces. Users provide an object class text $T_\text{o}$ and a grounding text $T_\text{g}$.
The objective of 3D zero-shot segmentation is to accurately segment the regions specified by the text within the given mesh.




Current 3D semantic segmentation models based on untextured meshes can only utilize shape information for segmentation, which is easily influenced by the lighting conditions and the viewpoints. We propose leveraging generated textures to incorporate texture information for mesh segmentation, rather than relying solely on shape information to address this issue. 
As shown in Fig. \ref{fig:method}, we propose a novel 3D mesh semantic segmentation framework, consisting of three parts, including text-guided texture synthesis, 2D zero-shot semantic segmentation, and face confidence revoting strategy.

In text-guided texture synthesis, we generate textures for untextured mesh $M_\text{unTextured}$ based on object class text $T_\text{o}$. The texture information comes from the pre-trained generative model Stable Diffusion \cite{rombach2022high} which has great prior texture information. We render the textured meshes $M_\text{Textured}$ and the untextured meshes $M_\text{unTextured}$ to 2D images in different viewpoints and segment the 2D images according to the object class $T_\text{o}$ and the grounding description $T_\text{g}$ by the 2D zero-shot segmentation model GroundingDINO \cite{liu2023grounding} and SAM \cite{kirillov2023segment}. Finally, we fuse the segmentation results from different views and obtained the segmented faces. Considering the inconsistency among the predictions from different views, especially from the extreme hard-case views, we further propose to fuse the 2D segments by the face confidence voting module and achieve consistency among different views.

\subsection{Text-Guided Texture Synthesis}
Our method segments the textured mesh and text-guided texture synthesis first generates textures given the raw untextured meshes guided by the text descriptions. 
As shown in Fig. \ref{fig:method}, the untextured meshes only provide the structure information. It is hard to localize the semantic regions without colors, for example, the car doors that are integrated with the car body structure.
Therefore, we leverage the texture information generated from the pre-trained generative model such as Stable Diffusion \cite{rombach2022high} to help the mesh segmentation. The generative model is trained on large-scale of data therefore is good at generating textures.

It should be noted that the descriptions used in this module are different from the grounding descriptions. For example, our model employs ``a photo of $\{T_{o}\}$ and $\{\}$ view'' as the prompt for the diffusion model, while the grounding descriptions are like ``hair'', ``door'' or ``tail''. 
TEXTure\cite{richardson2023texture} introduces different view prompts to facilitate better generation conditioned on viewpoints, which include ``front'', ``left'', ``back'', ``right''. Further details on the prompt settings for other orientations are available in the supplementary material. 

\begin{figure*}[!t]
\centering
\includegraphics[width=0.95\textwidth]{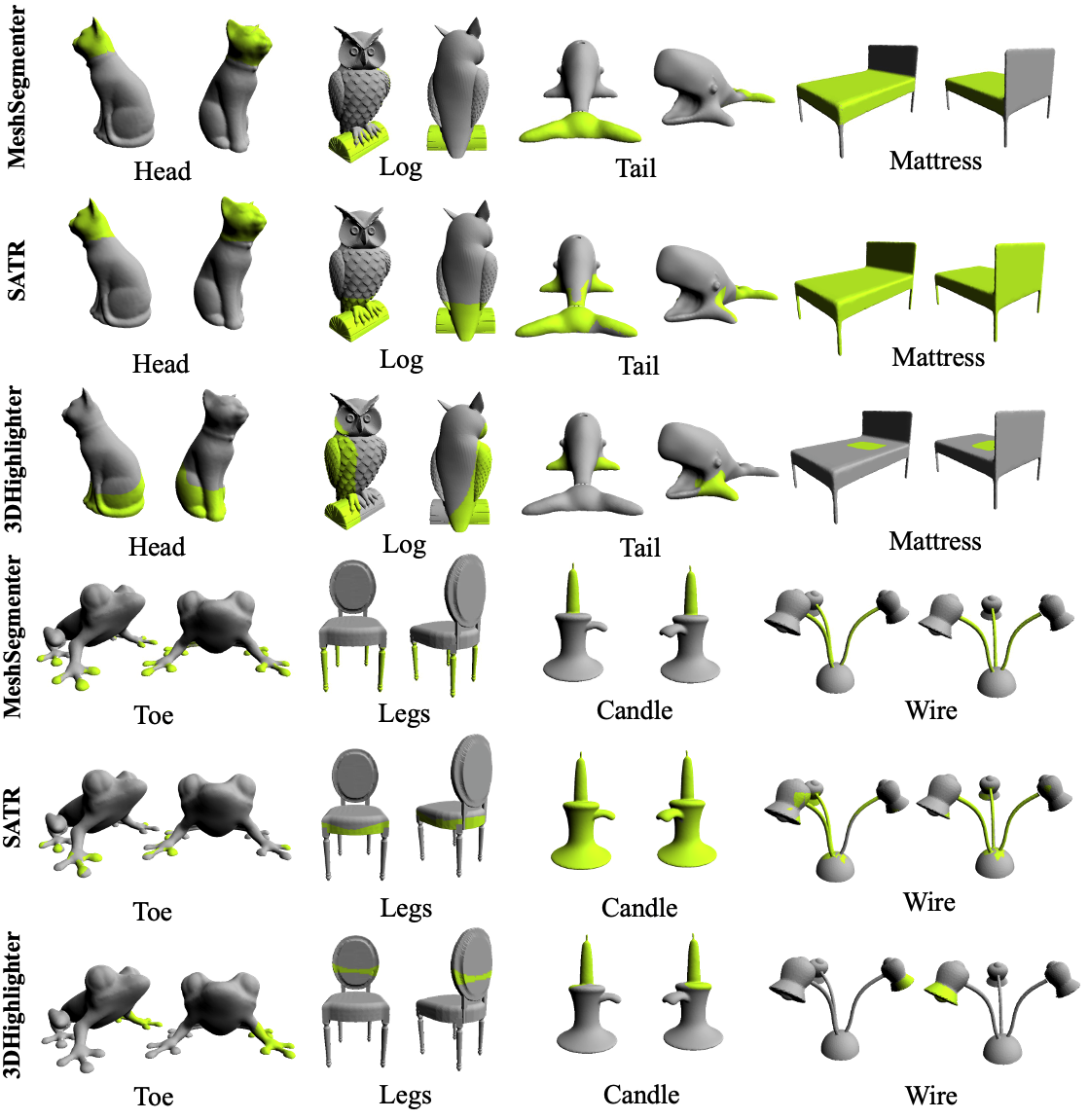}
\caption{Qualitative results in the single query: MeshSegmenter performs efficient zero-shot mesh semantic segmentation across diverse meshes.}
\label{fig:single_results}
\end{figure*}

\begin{figure}[h]
\centering
\includegraphics[width=\textwidth]{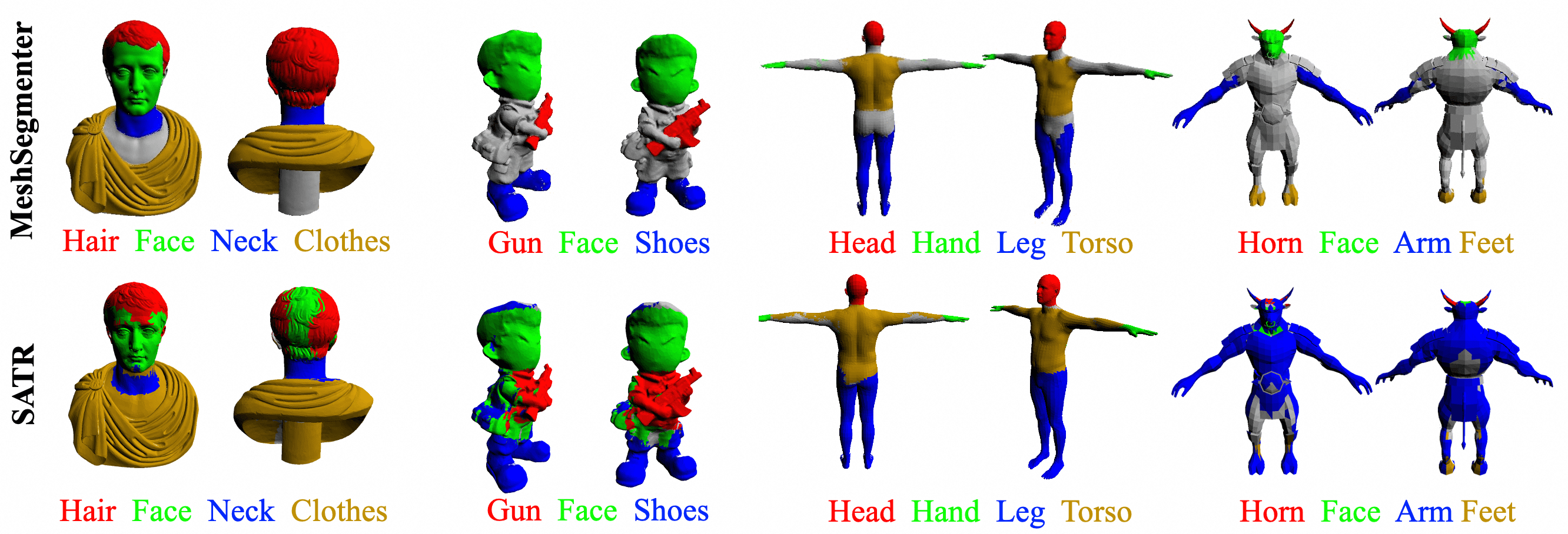}
\caption{Qualitative results in the multiple queries: MeshSegmenter performs accurate zero-shot mesh semantic segmentation, which does not rely on competition between multi queries segmentations. For instance, in the first column of MeshSegmenter, ``chest'' is not mistakenly segmented as ``clothes''. In the third column of MeshSegmenter, the ``buttocks'' are not mistakenly segmented as ``leg'' or ``torso''.}
\label{fig:multi_results}
\end{figure}

\subsection{2D zero-shot semantic segmentation}

Our work utilizes both $M_\text{unTextured}$ and $M_\text{Textured}$ meshes to provide geometric and texture information, respectively. The processing workflow is consistent for both types of meshes. For simplicity, we will explicitly enumerate only the segmentation steps for $M_\text{unTextured}$.

Get the rendered $K$ images, $x_k$ is the rendered image, and $\text{VisFace}_k$ is visible face set in the $k$-th viewpoint. 
Randomly generating viewpoints during rendering can have a detrimental effect on the efficiency of object semantic segmentation. This can result in the introduction of extreme perspectives, which can cause inaccuracies in semantic segmentation. To address this concern, we define a camera rendering trajectory that aims to strike a balance between the semantic segmentation results from rendered images and the coverage of all object viewpoints.
In particular, we first render $M_\text{unTextured}$ from $K$ viewpoints $V\in \{v_1,\dots,v_l\}$. We set radius $r$ as 2, and for the polar angle $\theta$ being $75 ^{\circ}$ and $115 ^{\circ}$ respectively, we have the azimuthal angle $\phi$ uniformly sampled $\frac{K}{2}$ times in a range of $0 ^{\circ}$ to $360 ^{\circ}$.
The rendering process is formulated as:
\begin{equation}
    \text{Render}(M_\text{unTextured}, v_k) = (x_k, \text{VisFace}_k).
\end{equation}

GroundingDINO\cite{liu2023grounding} is a novel approach to open-set object detection that combines the strengths of DINO architecture and grounded pre-training model. It takes an image $x_k$ and a 2D region-level description $T_\text{g}$ as input and accurately segments the region specified by the text into bounding boxes $b_k$ and corresponding 2D confidence scores $c_k$.
The text-guided detection process is formulated as:
\begin{equation}
    D (x_k, T_{g}) = (b_k, c_k ).
\end{equation}

Segment Anything Model (SAM) \cite{kirillov2023segment} is a foundational approach to image segmentation that enables promotable segmentation. This means the model can be trained on a pre-training task and then used to respond appropriately to any prompt at inference time, making it highly adaptable to a range of downstream tasks. An inference approach is to use the bounding box as input and SAM\cite{kirillov2023segment} outputs the mask $\text{Mask}_k$ for the region of interest within the bounding box $b_k$. 
The process is formulated as:
\begin{equation}
    \text{SAM}(b_k)=(\text{Mask}_k).
\end{equation}


Our model is specifically developed for performing zero-shot semantic segmentation of 3D objects, where the task involves delineating a region specified by text from an object. These regions are required to be partial regions of the object. Therefore, when detection produces bounding boxes that cover the entire object, we classify these as evident segmentation errors. These detection results are promptly deleted.




\begin{figure}[!t]
\centering
\includegraphics[width=0.85\textwidth]{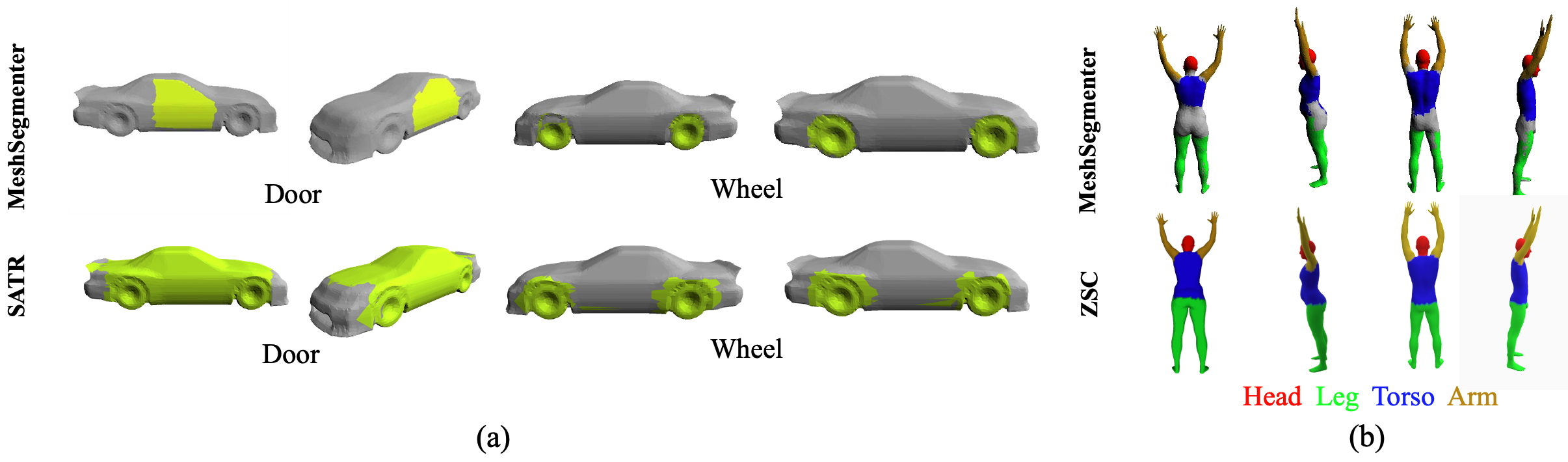}

\caption{
(a): Semantic Segmentation on Regions
with Weakly Geometric Signal.(b) Qualitative results with ZSC.
}
\label{fig:zsc}
\end{figure}


\subsection{Face Confidence Revoting (FCR)}

Based on the results of a powerful 2D detection and segmentation model, for each rendered image, we can obtain a set of bounding boxes $b_k$, 2D confidence scores $c_k$, and accurate masks $\text{Mask}_k$ for the regions described in text within that viewpoint. Additionally, using Nvidia Kaolin, we can obtain a face index map for that particular viewpoint. This allows us to identify visible faces $\text{VisFace}_k$ and determine whether they are within the mask region.

A simple approach would be to aggregate the faces of the mask region across different viewpoints to obtain the final segmentation result. However, this approach would introduce errors from incorrectly detected and segmented faces in a particular viewpoint, thus contaminating the overall segmentation result. Therefore, we propose a technique called Face Confidence Revoting.

In Face Confidence Revoting, we selectively consider the faces within the mask region with their corresponding confidence scores, while penalizing visible faces outside the mask region. This approach allows the correct segmentation results of neighboring faces to compensate for the errors in detection and segmentation within the current viewpoint. We denote the global confidence scores from $M_\text{unTextured}$ or $M_\text{Textured}$ as $g_m$
For the visable faces in k-th view $v_k$, we set the local confidence $l_k$ of the masked faces to $c_k$ and the unmasked faces to $-c_k$. The global confidence of the m-th face is the sum of multi-view local confidence:
\begin{equation}
    g_m = \sum_{i=1}^{k} l_{i}
\end{equation}


MeshSegmenter leverages the $M_\text{unTextured}$ and $M_\text{Textured}$ generated from the generative models to obtain the overall semantic segmentations. Consequently, our strategy involves setting the overall confidence scores $o_m$ as the average of the $g_m$ values of the untextured and textured meshes. When the value of $f_m$ exceeds $o_{\text{threshold}}$. It is considered to belong to the region determined as a text. 
\begin{equation}
\begin{cases}f_m \in R_{\text{text}}, & \text { if }o_m > o_{\text{threshold}}  \\
f_m \notin R_{\text{text}}, & \text { if }o_m \le o_{\text{threshold}} \\ \end{cases}
\end{equation}
$R_{\text{text}}$ is a collection of faces for a text-specified region.
We set the threshold as 0 for all cases to ensure a fair comparison. Finally, we averages the confidence of each face with their neighbor faces to obtain a smooth segmentation. 

\section{Experiment}

\subsection{Implementation Details}
We leveraged a single Nvidia A40 GPU for each experiment, involving a single object, its corresponding description, and a grounding text. 
We use Nvidia Kaolin to render multi-view 2D images from $M_\text{Textured}$ and $M_\text{unTextured}$. We compare MeshSegmenter with 3DHighlighter\cite{decatur20223d}, SATR\cite{abdelreheem2023satr}, and ZSC\cite{abdelreheem2023zero}. Since ZSC\cite{abdelreheem2023zero} is not open-source, we use its official results for comparison. We utilize existing 2D detection model GroundingDINO\cite{liu2023grounding} and segmentation model SAM\cite{kirillov2023segment}, selecting the segmentation results corresponding to the bounding box with the highest confidence score from different perspectives as our baseline.

\subsection{Zero-shot Mesh Semantic Segmentation}
\textbf{Qualitative Comparisons.}
MeshSegmenter demonstrates superior performance in zero-shot mesh semantic segmentation compared with 3DHighlighter\cite{decatur20223d} and SATR\cite{abdelreheem2023satr}, as shown in Fig. \ref{fig:single_results}. 
While 3DHighlighter\cite{decatur20223d} is limited to segmenting based on a single text query, MeshSegmenter can accurately segment a mesh in single and multiple queries. When dealing with multiple queries, MeshSegmenter does not rely on the competition of semantic segmentation for neighboring text queries and can independently and accurately segment each semantic region. We show the comparisons with SATR\cite{abdelreheem2023satr} in Fig. \ref{fig:multi_results} and the comparisons with ZSC \cite{abdelreheem2023zero} in Fig. \ref{fig:zsc} (b). Texture systhesis can effectively enhance the performance of zero-shot semantic segmentation, particularly for regions with minimal geometric signals, as shown in Fig. \ref{fig:zsc} (a).

\noindent\textbf{Quantitative Comparisons.} We evaluate the performance of 3D zero-shot segmentation using our proposed MeshSegmenter on the ShapeNetPart\cite{Yi_Kim_Ceylan_Shen_Yan_Su_Lu_Huang_Sheffer_Guibas_2016} dataset. The ShapeNetPart\cite{Yi_Kim_Ceylan_Shen_Yan_Su_Lu_Huang_Sheffer_Guibas_2016} dataset contains 31,963 objects and 50 annotated parts. Due to the computational resource-intensive nature of our model, which relies on multi-view texture generation, we randomly selected 20 objects from each category for evaluation. We compared our results with 3DHighlighter\cite{decatur20223d} and SATR\cite{abdelreheem2023satr}, and our approach clearly outperforms them, as shown in Tab. \ref{tab:results}.

\noindent\textbf{User Study.} We conducted a user study on our dataset of 20 3D meshes, with 30 respondents rating segmentation results on a scale of 1 to 5, with higher values indicating better performance. MeshSegmenter outperforms both 3DHighlighter\cite{decatur20223d} and SATR\cite{abdelreheem2023satr} in single query and multiple queries tasks, as shown in Tab. \ref{tab:user}.

\begin{table*}[!t]

\caption{Quantitative comparisons on PartNet subset}
\centering
\begin{tabular}{l|ccccc}
\midrule
Method        & mIou (\%)                    & Bag                     & Knife                   & Earphone              & Chair  \\
\midrule
3DHighlighter\cite{decatur20223d} & {5.8} & {2.4} & {3.2} & {7.8}&{9.7} \\
\midrule
SATR\cite{abdelreheem2023satr} & {43.4} & {50.2} & {53.7} & {32.5} & {37.2}\\
\midrule
MeshSegment-Baseline(Ours)    & 42.3                    & 47.1                    & 38.2                    & 32.2      & 51.7              \\
\midrule
MeshSegmenter(Ours) & \textbf{69.0}                    & \textbf{78.3}                 & \textbf{68.2}                   & \textbf{57.2}   & \textbf{72.2}\\
\midrule
\end{tabular}
\label{tab:results}

\end{table*}

\begin{table*}[!t]

\caption{User Study. MeshSegmenter demonstrates superior performance in both single query and multiple queries tasks.}
\centering
\begin{tabular}{l|cc}
\midrule
Method        & Single                  & Multiple                       \\
\midrule
3DHighlighter\cite{decatur20223d} & {5.8} & {2.4}  \\
\midrule
SATR\cite{abdelreheem2023satr} & {43.4} & {50.2} \\
\midrule

MeshSegmenter(Ours) & \textbf{69.0}  & \textbf{78.3}    \\
\midrule
\end{tabular}
\label{tab:user}
\end{table*}

\begin{figure}[!t]
  \centering
  \begin{minipage}{0.5\textwidth}
     \centering
     \includegraphics[width=\textwidth]{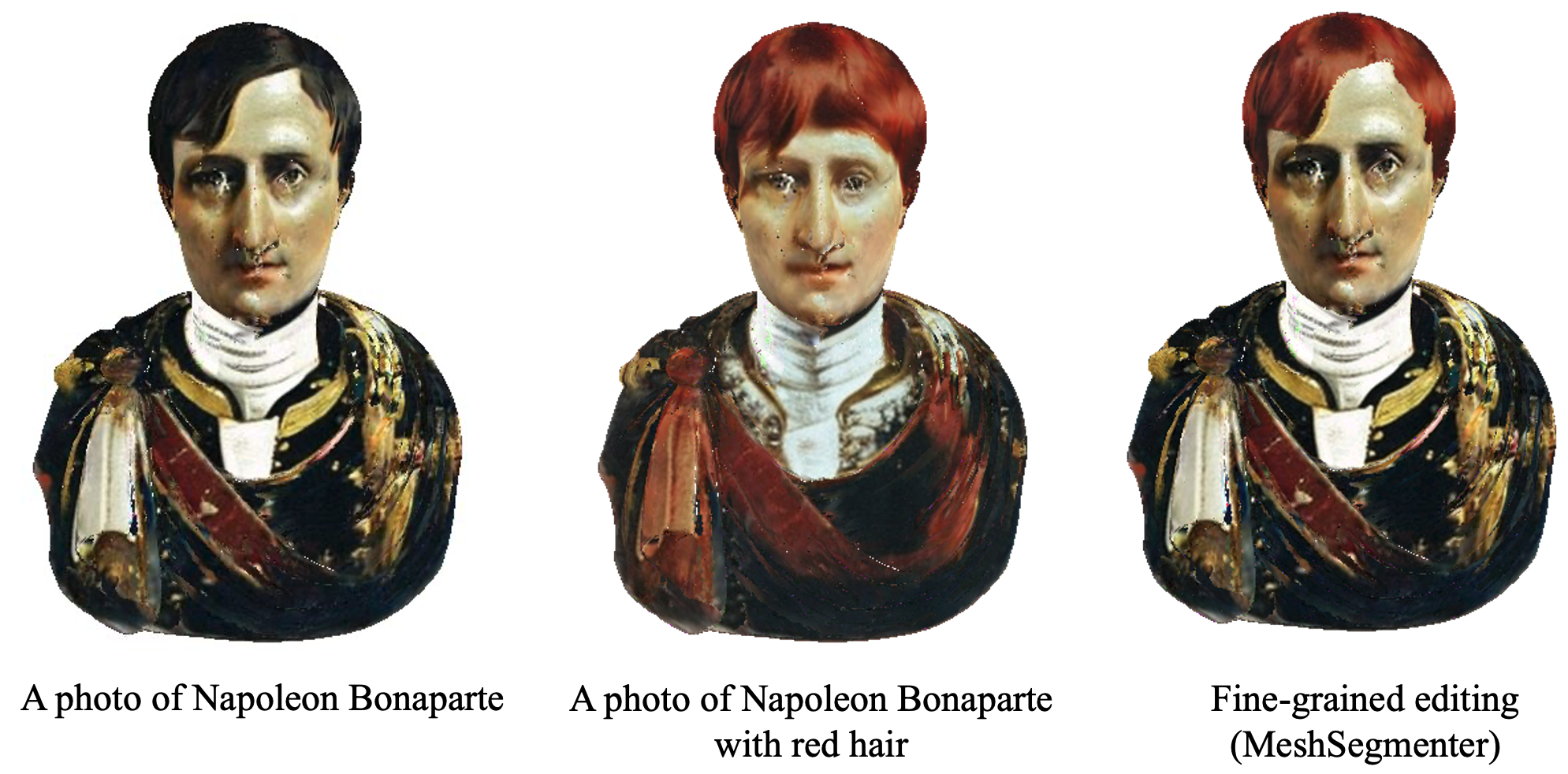} 
    
     \label{fig:image}
  \end{minipage}%
  \begin{minipage}{0.5\textwidth}
     \centering
    \caption{Controlled Mesh Stylization. 
MeshSegmenter can accurately segment regions specified by text, significantly enhancing the capability for fine-grained editing. For example, it precisely segments editable region, allowing for detailed modifications such as changing black hair to red.}
\label{fig:style}
  \end{minipage}
\end{figure}



\subsection{Application of MeshSegmenter}

\noindent\textbf{Controlled Mesh Stylization}
Previous mesh stylization work focuses solely on the stylization of the entire object. Local descriptive prompts could easily corrupt the global stylization appearance. It is challenging to accurately stylize the text-specified region while preserving other areas from being corrupted. MeshSegmenter can effectively locate text-specified local regions to achieve controllable local editing, as shown in Fig. \ref{fig:style}. We begin with a human mesh and stylize it using the prompt ``a photo of Napoleon Bonaparte''. For a fair comparison, we use the same seed to stylize the same mesh with the prompt ``a photo of Napoleon Bonaparte with red hair''. TEXTure\cite{richardson2023texture} noticeably corrupts the ``ribbon'' region with red information, which is not specified by the grounding text ``hair''. In contrast, MeshSegmenter effectively segments the ``hair'' region, enabling precise mesh editing.

\begin{figure}[!t]
\includegraphics[width=\textwidth]{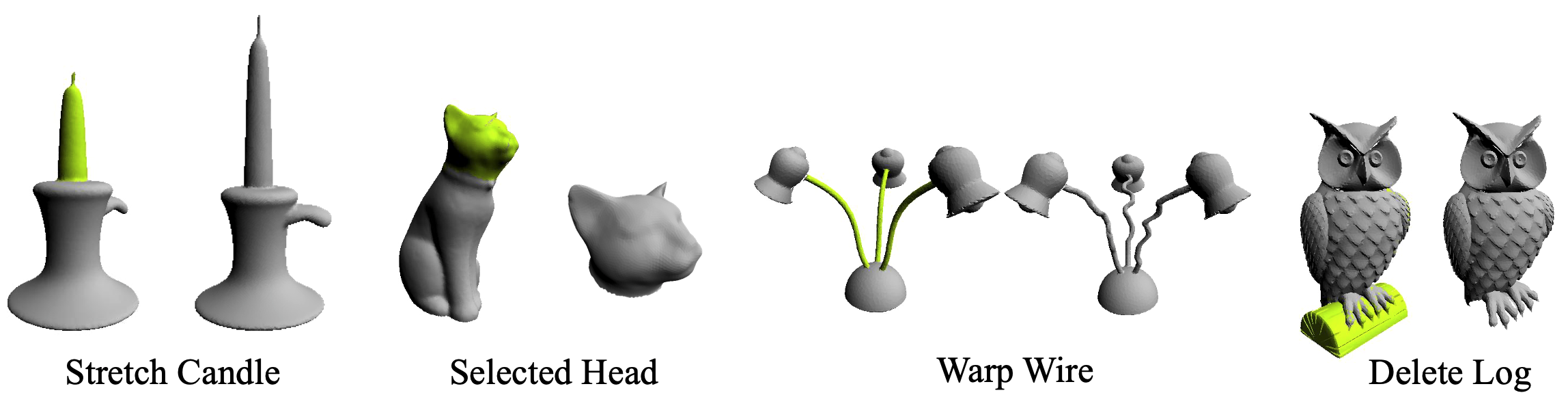}

\caption{
Mesh Editing: Utilizing the accurate zero-shot mesh semantic segmentation results provided by MeshSegmenter, we can perform precise mesh editing tasks such as stretching, selecting, warping, and deleting  text-specified regions.
}
\label{fig:Selective}
\end{figure}

\noindent\textbf{Mesh Editing} Currently, mesh editing mainly focuses on object-level editing, such as adding, deleting, and scaling. Precisely segmenting the text-specified region can bring about fine and powerful object script editing capabilities. Our proposed model has demonstrated the robust semantic segmentation capability of 3D objects through a combination of object description and grounding text. This approach can be readily integrated into 3D rendering applications. We can input diverse 3D object meshes and leverage object grounding to select the segmentation region based on the text input automatically. Coupled with the mesh editing functionalities of 3D rendering applications, we can precisely manipulate the region corresponding to the grounding text, thus providing direct control over the editing process without requiring challenging manual selection, particularly for non-uniform regions, as shown in Fig. \ref{fig:Selective}.

\begin{figure}[t]
\centering
\includegraphics[width=0.5\textwidth]{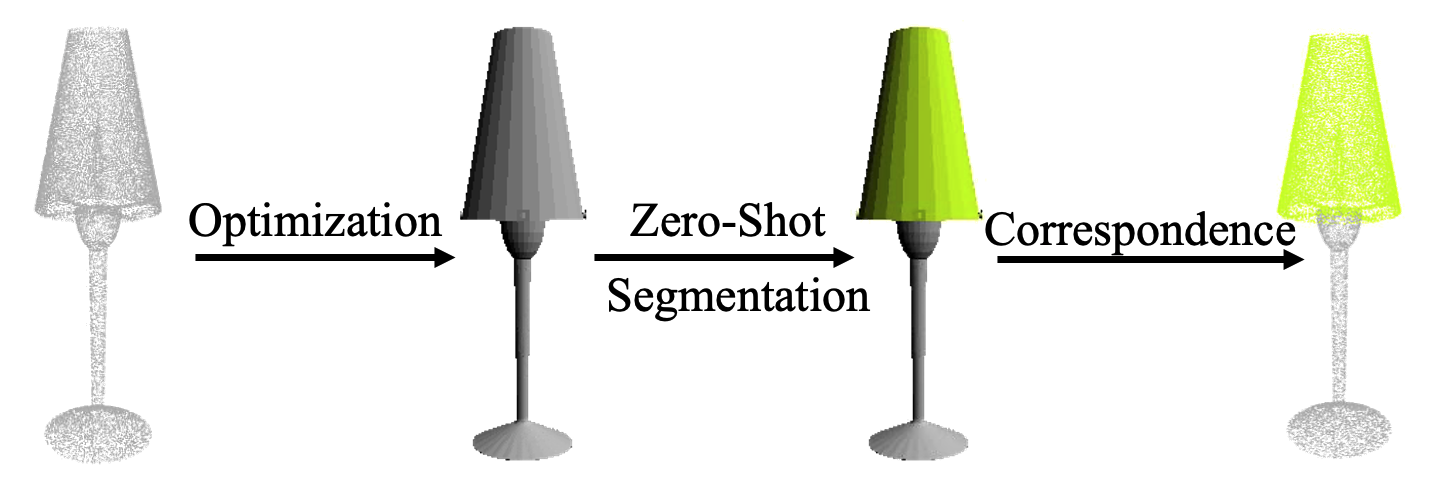}
\caption{Point Cloud Semantic Segmentation: 
We optimize a mesh result to obtain the occlusion relationship of 3D data using \cite{peng2021shape} and finally correspond the mesh segmentation result to the point cloud. Our model can also segment point cloud data effectively.}

\label{fig:pointcloud}

\end{figure}

\noindent\textbf{More 3D Representations}
MeshSegmenter not only performs precise and effective semantic segmentation on mesh structures, but this capability can also be extended to other 3D representations. MeshSegmenter relies on the detection and segmentation results of images rendered from various viewpoints of a 3D object, as well as the mapping relationship between 2D segmentation regions and the 3D object. Point clouds can accurately represent a variety of 3D objects but fail to convey their topological structure, making it challenging to establish the mapping relationship between rendered images and point clouds, and hence difficult to determine the visibility of point clouds from specific viewpoints. Therefore, we optimize point clouds into mesh structures to capture the object's topological structure, and then align the zero-shot 3D semantic segmentation results on the mesh to the point cloud structure, as shown in Fig. \ref{fig:pointcloud}. MeshSegmenter can be extended to more 3D representations when the mapping relationship between 2D rendered images and 3D representations is obtained.




\begin{table}[!t]
\caption{Ablation studies of different modules on PartNet subset}
\centering
\begin{tabular}{lll|c}
\midrule
{Shape-branch} & Texture-branch & Face Revoting & mIoU (\%) \\ \midrule
  \quad\quad\checkmark                    &                &               & 42.3 \\ 
\quad\quad\checkmark  &    \quad  \quad   \checkmark            &               & 49.7 \\ 
     \quad  \quad   \checkmark                 &               &     \quad  \quad   \checkmark           & 52.5   \\ 
                      &       \quad  \quad   \checkmark         &   \quad  \quad   \checkmark            & 62.2 \\ 
      \quad  \quad   \checkmark                &      \quad  \quad   \checkmark          &       \quad  \quad   \checkmark        & \textbf{69.0}   \\ \midrule
\end{tabular}
\label{tab:ablation}

\end{table}



\begin{figure}[!t]
\centering
\begin{minipage}[t]{0.48\textwidth}
\centering
\includegraphics[width=6cm]{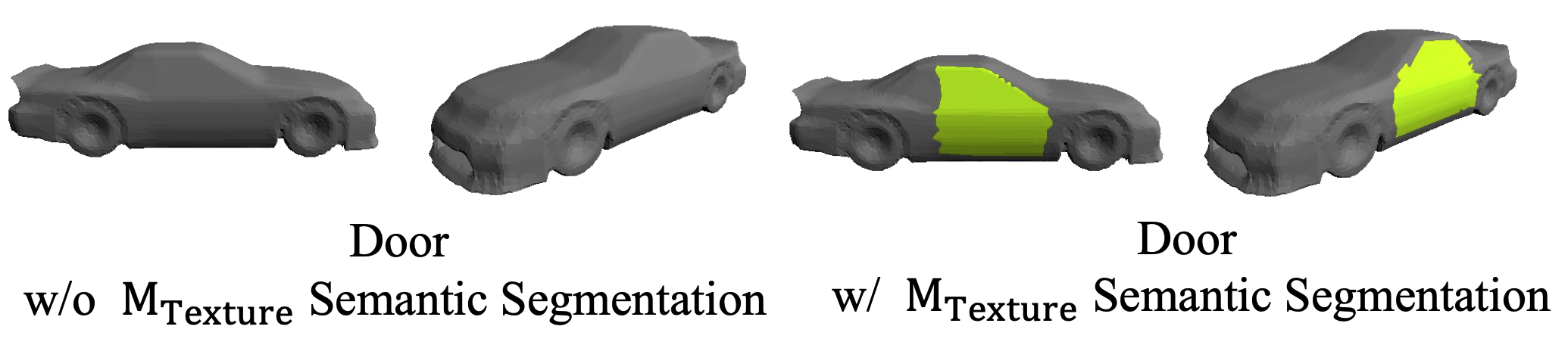}
\caption{Ablation Study of Segmentation via Texture Synthesis. }
\label{fig:ablation_texture}
\end{minipage}
\begin{minipage}[t]{0.48\textwidth}
\centering
\includegraphics[width=6cm]{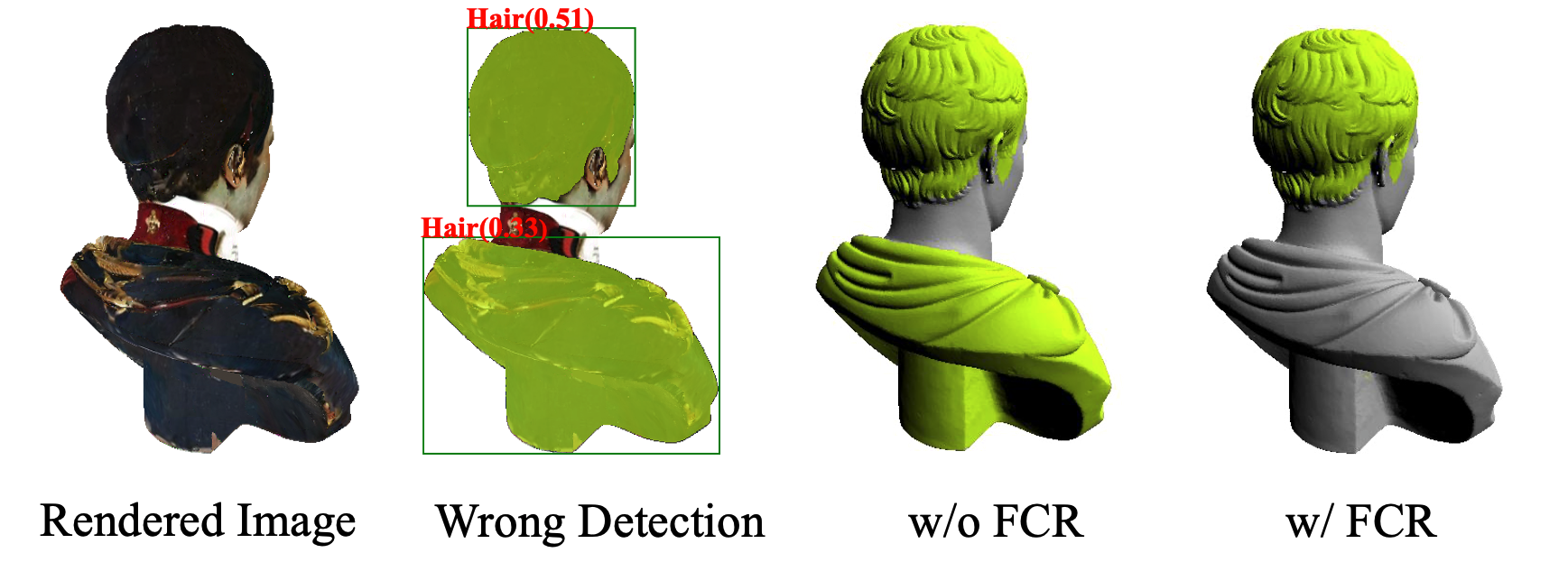}
\caption{Ablation Studies of Face Confidence Revoting.}
\label{fig:ablation_revting}
\end{minipage}
\end{figure}

\subsection{Ablation Studies}

\noindent\textbf{Effectiveness of the Segmentation via Texture Synthesis}
Both the 2D zero-shot detection model and segmentation model are trained on textured images, and the application of texture synthesis for segmentation effectively reduces the domain gap between inference images and training images. Additionally, regions with insignificant geometric information are difficult to segment without textured meshes. Extracting texture information from the generative model allows for effectively segmenting the text-specified region, as shown in Fig. \ref{fig:ablation_texture}. 

\noindent\textbf{Effectiveness of the Face Confidence Revoting.}
We ablate the effectiveness of our proposed Face Confidence Revoting, as shown in Fig. \ref{fig:ablation_revting}. The most direct approach to aggregating multi-view detection and segmentation results is to assign each detected face to the text-specified region with the highest confidence. However, this approach will introduce incorrect local detection and segmentation results into the global results. To address this issue, MeshSegmenter assigns negative local confidence scores to faces that are not within the segmented regions and aggregates local confidence scores into a global confidence score, using neighboring viewpoints to correct erroneous segmentation results.

\begin{figure}[!t]
  \centering
  \begin{minipage}{0.5\textwidth}
     \centering
     \includegraphics[width=\textwidth]{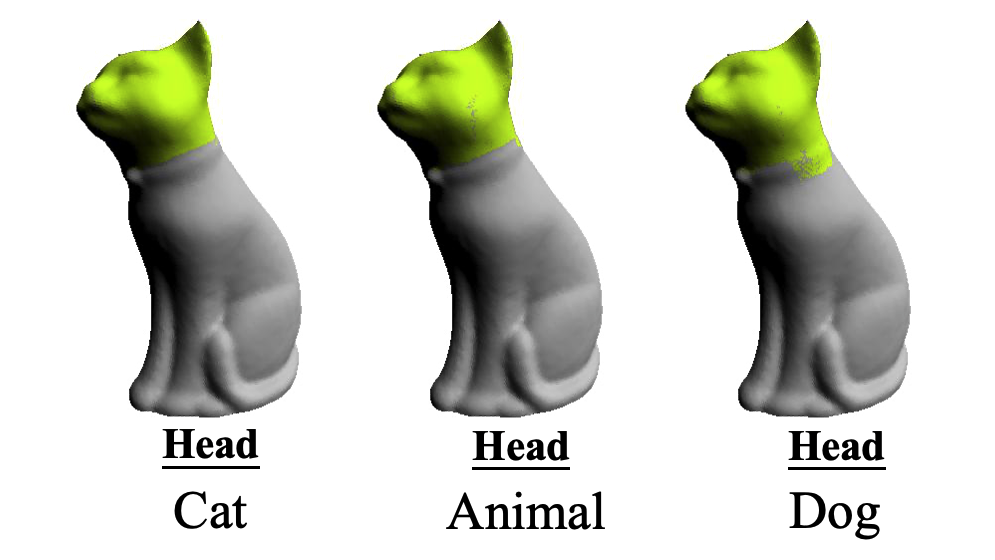} 
    
     \label{fig:image}
  \end{minipage}%
  \begin{minipage}{0.5\textwidth}
     \centering
    \caption{Robustness of Object Description: MeshSegmenter exhibits strong robustness towards diverse object descriptions. It not only achieves accurate results for the correct description of objects ``cat'' with the grounding description ``head'', but it also achieves precise segmentation for homogeneous description of objects, such as animals and dogs.}
\label{fig:class}
  \end{minipage}
\end{figure}

\noindent\textbf{Robustness of Object Description} The 3D zero-shot segmentation task should maintain robustness for diverse object descriptions. MeshSegmenter effectively achieves 3D zero-shot segmentation by leveraging geometric and color information of object meshes through text. Accurate description of an object's class can lead to the most appropriate texture generation, however, this poses a challenge for users, particularly when confronted with complex objects. Nonetheless, MeshSegmenter is still able to obtain effective texture information through inaccurate object descriptions and achieve accurate and robust segmentation, as shown in Fig. \ref{fig:class}.
\noindent\textbf{}



\section{Limitation}

 MeshSegmenter effectively aggregates detection and segmentation results from both textured and untextured meshes across multiple viewpoints, achieving accurate semantic segmentation for text-specified regions. However, large-scale pre-trained 2D models are not designed for local-level detection and segmentation, which can lead to inaccuracies, including the erroneous detection of entire objects as target regions. Our model addresses this challenge by filtering out detections that encompass whole objects and introducing the face confidence revoting strategy. This approach effectively corrects erroneous detections based on correct detections from adjacent viewpoints. Another issue to consider is the visibility of faces. Our method infers 3D semantic segmentation from multi-view 2D detection and segmentation results. However, no rendering strategy can ensure the visibility of every face. Our approach to sampling rendering viewpoints involves uniform sampling rendered viewpoints in fixed rotation, which maximizes the likelihood that more faces will be visible.

\section{Conclusion}

In this work, we have proposed MeshSegmenter, a pioneering framework for 3D zero-shot semantic segmentation.
Our model extends the capabilities of 2D detection and segmentation models to accurately segment diverse 3D meshes by text.
We propose generating textures based on object descriptions to enhance the accuracy of 2D segmentation models. We also develop a multi-view revoting module that ensures 3D consistency in segmentation results.
MeshSegmenter provides a transformative tool for 3D zero-shot segmentation, with potential applications in computer graphics and computer vision. By leveraging 2D segmentation models, incorporating texture information, and integrating multiple views, our framework provides stable and reliable segmentation results.

\section*{Acknowledge}
The work was supported by NSFC $\#61932020$, $\#62172279$ and Program of Shanghai Academic Research Leader.


%
%

\end{document}